\documentclass[letterpaper, 10 pt, conference]{ieeeconf}  % Comment this line out if you need a4paper
\IEEEoverridecommandlockouts                              % This command is only needed if
\usepackage{graphics} % for pdf, bitmapped graphics files
\usepackage{epsfig} % for postscript graphics files
\usepackage{mathptmx} % assumes new font selection scheme installed
\usepackage{times} % assumes new font selection scheme installed
\usepackage{amsmath} % assumes amsmath package installed
\usepackage{amssymb}  % assumes amsmath package installed
\usepackage{mathtools}
\usepackage{subfigure}
\usepackage{epstopdf}
\usepackage{color}
\usepackage{anyfontsize}
\usepackage{stfloats}
\usepackage{multirow}
\usepackage{soul}
\usepackage{algorithm}
\usepackage{adjustbox}
\usepackage{algpseudocode}
\usepackage{algorithmicx}
\usepackage{lipsum}
\usepackage{array}
\usepackage{url}
\usepackage{newtxtext, newtxmath}

\makeatletter
\newcommand{\algmargin}{\the\ALG@thistlm}
\makeatother
\newlength{\whilewidth}
\algdef{SE}[parWHILE]{parWhile}{EndparWhile}[1]
     {\parbox[t]{\dimexpr\linewidth-\algmargin}{%
        \hangindent\whilewidth\strut\algorithmicwhile\ #1\ \algorithmicdo\strut}}{\algorithmicend\ \algorithmicwhile}%
\algnewcommand{\parState}[1]{\State%
     \parbox[t]{\dimexpr\linewidth-\algmargin}{\strut #1\strut}}

\title{\LARGE \bf Visuo-Tactile Manipulation Planning Using Reinforcement Learning with Affordance Representation
}
\author{Wenyu Liang$^{1*}$, Fen Fang$^{1*}$, Cihan Acar$^{1}$, Wei Qi Toh$^{2}$, Ying Sun$^{1}$, Qianli Xu$^{1}$, and Yan Wu$^{1,2}$ %what is the concern here? will have a bit of hard time to say why co-first, but not in alphabetical order.
\thanks{This work is supported by the Agency for Science, Technology and Research (A\*STAR) under its AME Programmatic Funding Scheme (Project \#A18A2b0046).$^{*}$ Author with equal contribution.}% <-this % stops a space
\thanks{$^{1}$ A*STAR Institute for Infocomm Research (I$^2$R), Singapore. } %{\tt\small <>@i2r.a-star.edu.sg}. A and B contribute equally in this work.
\thanks{$^{2}$ A*STAR Institute of High Performance Computing (IHPC), Singapore.}%{\tt\small <>@ihpc.a-star.edu.sg}
}

\begin{document}
\thispagestyle{empty}
\pagestyle{empty}
\maketitle
\begin{abstract}
Robots are increasingly expected to manipulate objects in ever more unstructured environments where the object properties have high perceptual uncertainty from any single sensory modality. This directly impacts successful object manipulation. In this work, we propose a reinforcement learning-based motion planning framework for object manipulation which makes use of both on-the-fly multisensory feedback and a learned attention-guided deep affordance model as perceptual states. The affordance model is learned from multiple sensory modalities, including vision and touch (tactile and force/torque), which is designed to predict and indicate the manipulable regions of multiple affordances (i.e., graspability and pushability) for objects with similar appearances but different intrinsic properties (e.g., mass distribution). A DQN-based deep reinforcement learning algorithm is then trained to select the optimal action for successful object manipulation. To validate the performance of the proposed framework, our method is evaluated and benchmarked using both an open dataset and our collected dataset. The results show that the proposed method and overall framework outperform existing methods and achieve better accuracy and higher efficiency.
\end{abstract}

\section{Introduction}
Robots are increasingly used in various industrial applications (such as assembly \cite{nicolas2021towards}, logistics \cite{krug2016next}, warehouse automation \cite{correll2016analysis}, etc.) to improve the level of automation and efficiency as well as to reduce manpower cost and accident rates. Dense box packing or packaging is one of the challenging tasks for robotic systems \cite{dong2019tactile}, where complex actions are required to manipulate yet semi-known objects. A robot is required to plan and execute a series of actions to stow these objects firmly into a confined space for space optimization \cite{Wang2021dense}.

\begin{figure}[!t]
    \centering\vspace{0.2cm}
    \includegraphics[width=1.0\columnwidth]{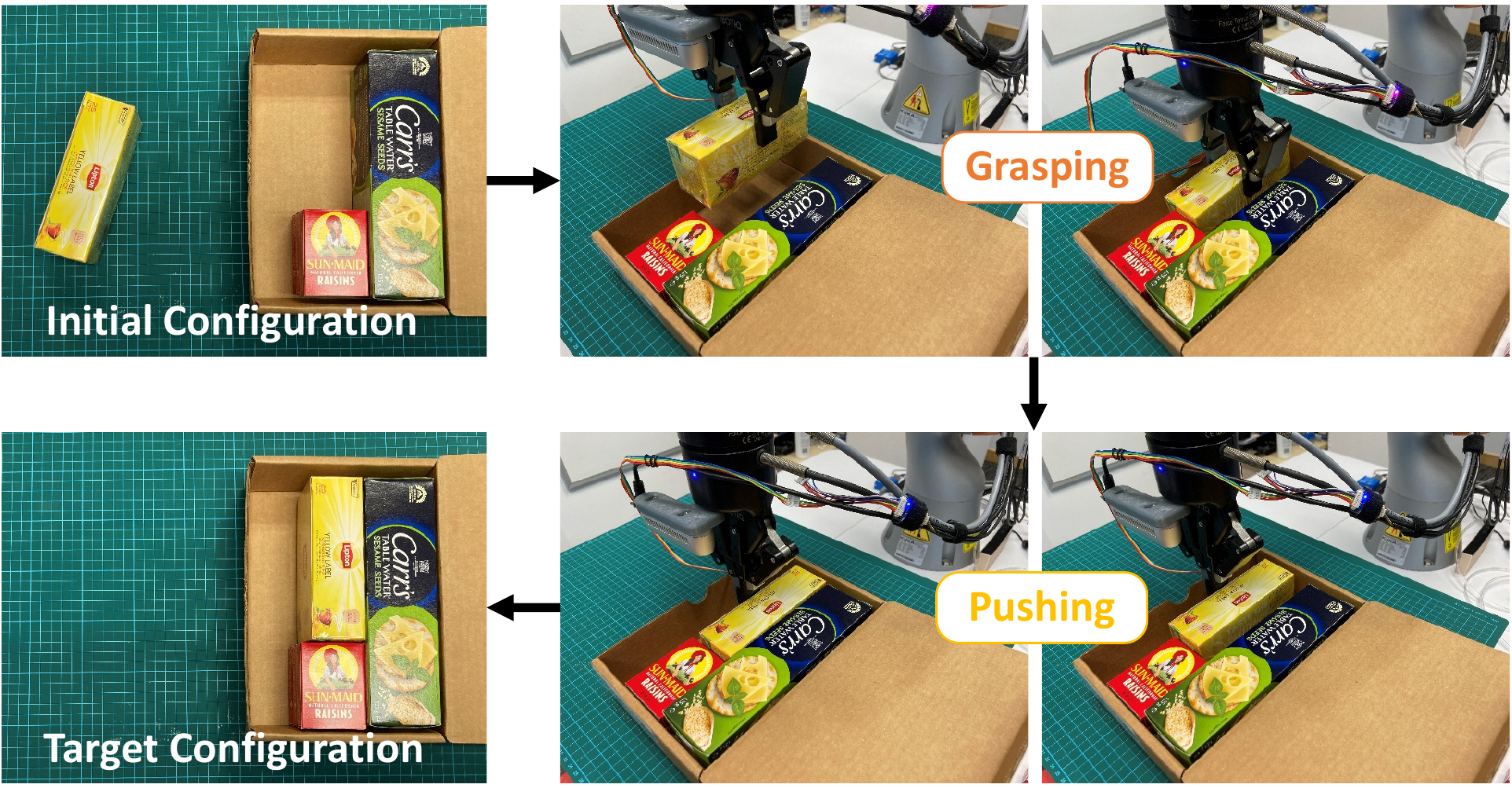}\vspace{-0.3cm}
    \caption{Overview of the object manipulation challenge}
    \label{fig:problem}\vspace{-0.5cm}
\end{figure}

Generally, the object packing task is composed of two key actions, pick-and-place and push. While pick-and-place constitutes the bulk of the action space, grasping action is often not sufficient to accomplish the dense packing task due to geometrical constraints (as illustrated in Fig. \ref{fig:problem}) and/or limited graspability of the object. In such cases, pushing action can help to move an object to the target location or to align with one another. It is therefore beneficial for the robot to be equipped with the capability to localise on the objects for optimal manipulation results. Such locations can be inferred using affordance learning approach.

%Why affordance?
Affordance, first introduced by James Gibson \cite{Gibson1966senses}, is a concept from psychology that describes the possibilities of an agent performing actions on an object \cite{ijcai2021-590}. In recent years, many researchers report the use of affordance in robotic manipulation. For example, the affordance models are successfully used in performing object positioning actions in a 2D space \cite{Hermans2013decoupling}, contact point selection for vision-based planar pushing \cite{kloss2020accurate}, vision-based robotic object pushing and grasping \cite{Wu2020learning}, dexterous robotic grasping \cite{Mandikal2021learning}. However, while affordances, especially visual ones, have been shown to accomplish a range of difficult object manipulation tasks, little attention is placed on inferring multiple affordances using a unified model \cite{Wu2020learning, andy0robotic} is less scalable for more complex motion planning and control applications.

%In summary, affordance can benefit the robots to accomplish object manipulation tasks.

%%%
%Why tactile?
Significantly, successful grasp and planar push can be challenging in a semi-structured environment by using visual input alone. While objects could appear visually similar, they may have different intrinsic properties (such as weight, center of mass (CoM), mass/density distribution, surface properties, etc.) making it difficult to be modelled solely through vision. Moreover, the difference in interactions between the end-effector and the objects is also hard to observe visually without significant changes. These in turn affect the accurate inference of the graspable/manipulable regions.

To tackle this challenge, the sense of touch can be used as a complementary sensory modality with vision. In \cite{Matthewaffordanceprediction2020}, Veres \emph{et al.} incorporates force/torque sensing into a deep grasp affordance prediction model for learning grasp affordance. Although it demonstrates enhancement in the success rate of the suction-based object grasping task, the force/torque sensor only provides single-point information which may not be sufficient for affordance modelling for contact-rich manipulation such as push. Tactile sensing, on the other hand, provides rich touch information that can be used to obtain a wide range of object-level information such as localization, mass, shape, size, CoM, slips, etc. \cite{Li2020review}.  In \cite{Hogan2018tactile}, Hogan \emph{et al.} introduces a tactile-based grasp policy as a regrasping stage after performing visual-affordance-based grasping to improve the robustness. However, tactile information is only used as a corrective/reactive signal outside the affordance model. 

%In \cite{kolamuri2021improving},  Kolamuri \emph{et al.} proposed a model-based algorithm using tactile sensors to detect the rotational displacement during grasping, which is integrated into a closed-loop regrasping framework for failure detection and stable regrasping. This research \cite{kolamuri2021improving} shows the effectiveness of using tactile sensor in stable grasping. 

%%
%%from hogan's paper, they said "to further improve grasping performance, we are interested in extending the idea of planning grasp adjustments by efficiently combining tactile and visual information" %%
%However, without integrating affordance representation of both modalities, regrasp is still required for learnt objects.
%%

%Why attention and RL?
It is also observed from the results of \cite{Matthewaffordanceprediction2020} that the learning model works better for objects with CoM closer to their visual centroids. We hypothesise that this is due to the dominance of the visual percepts. Hence, adding an attention mechanism \cite{vaswani2017attention} in the learning model will improve the performance of the affordance inference. This will also help to reduce the model's dependency on the quality of the support examples. Moreover, although the model presented in \cite{Matthewaffordanceprediction2020} can achieve high prediction accuracy, it requires a handful of explorations to guarantee such accuracy, and the number is fixed. To improve the manipulation efficiency, a method is required to reduce the number of explorations. To this end, a reinforcement learning-based (RL-based) motion planning algorithm can be used to close the perception-action loop and guide the robot to select an optimal action for the next exploration \cite{liang2021dexterous}. %Here, the use of RL is because it offers a powerful machine learning paradigm to solve for the optimal solution \cite{liang2021dexterous}.

%%%

In this work, a multisensory object manipulation planning framework integrating the affordance model into the reinforcement learning pipeline is proposed. The main contributions of this paper include: (i) A manipulation planning framework with multisensory feedback that outperforms on successful object manipulations; (ii) A unified deep multi-affordance representation learning model with built-in attention mechanism to effectively encapsulate both vision and multimodal touch information of an object as affordance map; (iii) The integration of affordance representation into a deep reinforcement learning pipeline to improve the efficiency and accuracy for planning the manipulation motions.

The rest of this paper is organized as follows. Section II describes the related affordance definitions as well as the robotic system used in this paper. Next, in Section III, the proposed framework design is presented in detail. Then, experiments on two different datasets are conducted and the results are discussed in Section IV. Finally, the conclusions are drawn in Section V.

\section{Definitions and System Description}
\label{sec:Background}
%To represent different objects, plastic building blocks with different configurations are used in this paper. 
The definitions of the affordances for graspability and pushabilities of the objects are explained as follows. 
%(e.g., shapes, mass distributions)

\begin{figure}[!t]
    \centering\vspace{0.2cm}
    (a) \subfigure{\includegraphics[width=0.85\columnwidth]{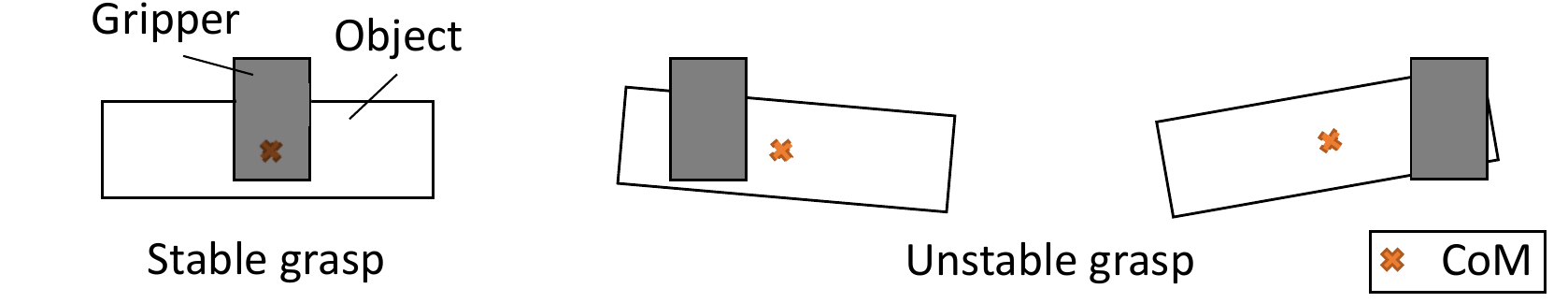}}\\\vspace{-0.2cm}
    (b) \subfigure{\includegraphics[width=0.85\columnwidth]{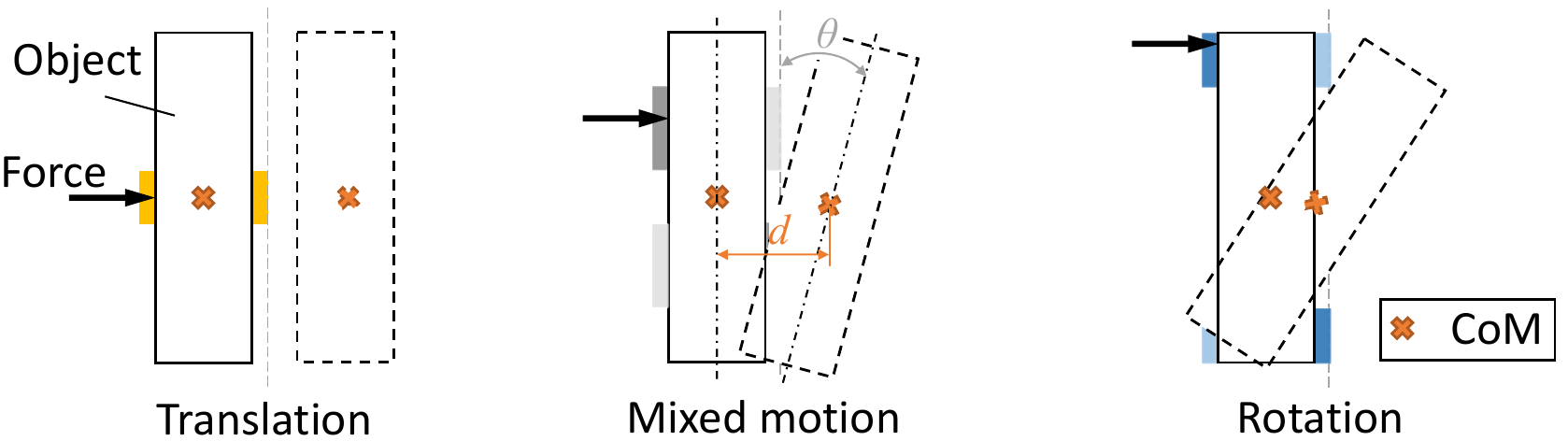}}\\
  \vspace{-0.3cm}
    \caption{Illustrations of the definition of the affordances for graspability and pushabilities: (a) graspability; (b) pushabilities: translation, rotation, and mixed motion (color zones denote the different pushable areas).}\vspace{-0.2cm}
    \label{fig:affordance}
\end{figure}

\noindent {{\textbf{Graspability:}}} A stable grasp is defined as graspable (see Fig. \ref{fig:affordance}(a)).

\noindent {{\textbf{Pushability:}}} A set of three lower-level affordances is defined to describe the affordability of an object when it is pushed: (i) translational pushability; (ii) clockwise (CW) rotational pushability; and (iii) counter-clockwise (CCW) rotational pushability. Figure \ref{fig:affordance}(b) illustrates pushability of an object.
%An illustration of pushability is shown in Fig. \ref{fig:affordance}(b).
%  \begin{itemize}
 % \item 
 
 \noindent\emph{- Translational Pushability:} A translational movement along the direction of a given push with consistent orientation is defined as translationally pushable (top of Fig. \ref{fig:affordance}(b)). The pushable zone (highlighted in yellow) is defined as the pushable surface of the object that only causes translation.
 %\item 
 
 \noindent\emph{- Rotational Pushability}: Although areas to both sides of the translationally pushable zone seem to be rotationally pushable, pushing some of these areas may result in both translation and rotation of the object (middle of Fig. \ref{fig:affordance}(b)). As such, a refined definition is needed. When rotational movement is caused by a given push of distance $d$, if the translation distance of the object's CoM does not exceed a threshold $d_{th}$ ($d_{th}=\beta d$, where $\beta < 1$ is a parameter to determine the threshold), the object is deemed rotationally pushable (bottom of Fig. \ref{fig:affordance}(b)). Qualified areas to the left and right of the translationally pushable zone are defined as CW and CCW rotationally pushable zones (highlighted in dark blue for CW and light blue for CCW), respectively.
%\end{itemize}

\begin{figure}[!t]
    \centering
   \includegraphics[width=0.9\columnwidth,trim = 0cm 0.05cm 0cm 0.25cm, clip=true]{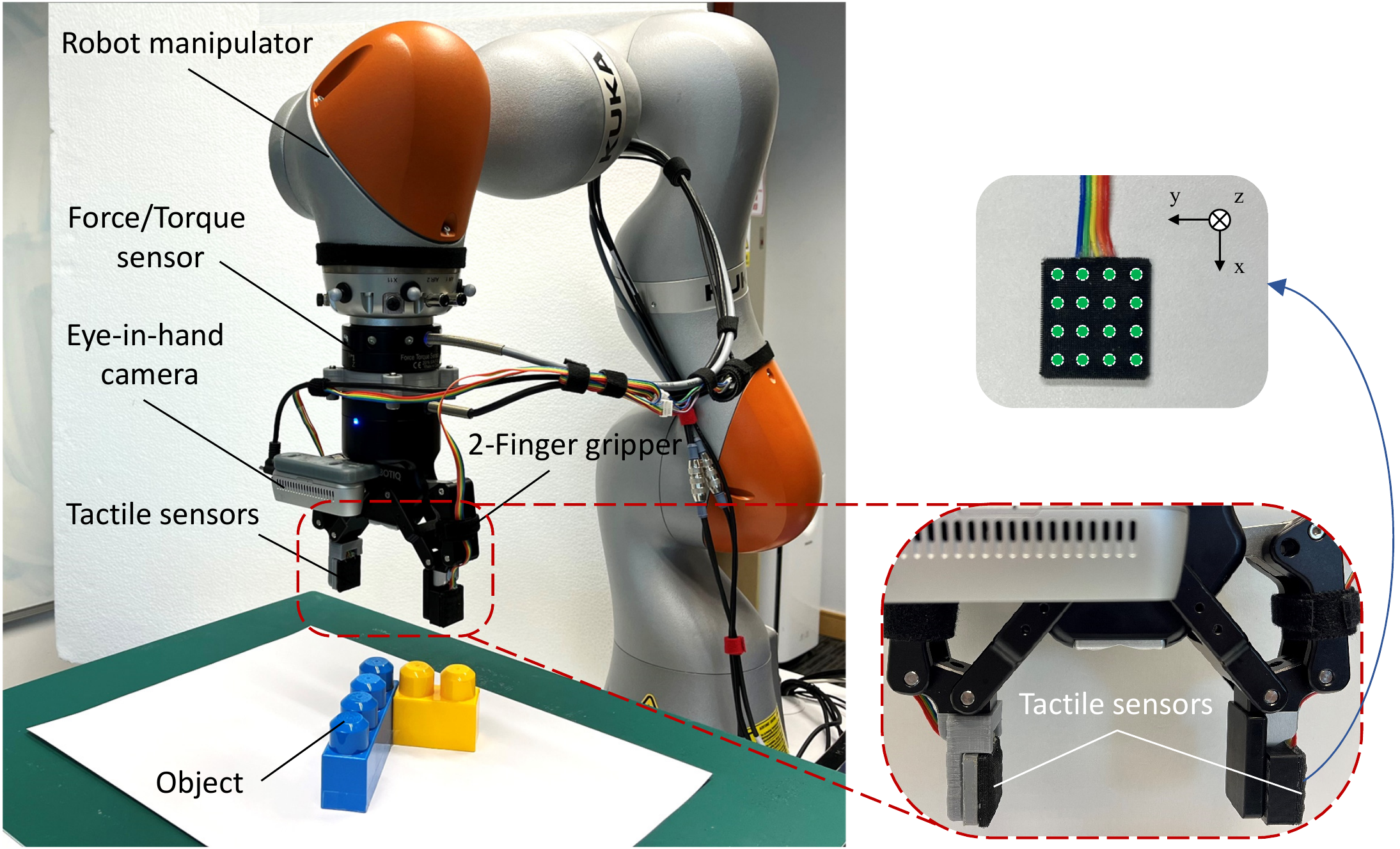}\\
      \vspace{-0.3cm}
    \caption{Robotic system setup.}
    \label{fig:robot}\vspace{-0.55cm}
\end{figure}

Figure \ref{fig:robot} shows the robotic system used in this work. This system consists of a KUKA LBR iiwa 14 R820 7-degrees-of-freedom (7-DoF) manipulator, a RealSense D435 eye-in-hand camera, a Robotiq FT 300 force/torque sensor (6-Axis measurement), a Robotiq 2F-85 2-finger gripper, and two XELA uSKin XR1944 tactile sensors mounted on the gripper's fingers. The tactile sensor has 16 taxels arranged in a 4-by-4 array and each taxel can detect the applied forces in 3D space (i.e., totally 48 outputs form the tactile sensor). Besides that, to represent different objects, plastic building blocks with different configurations are used.

\section{Framework Design}
\begin{figure}[!t]
  \centering\vspace{0.2cm}
  \centerline{\includegraphics[width=1.0\columnwidth]{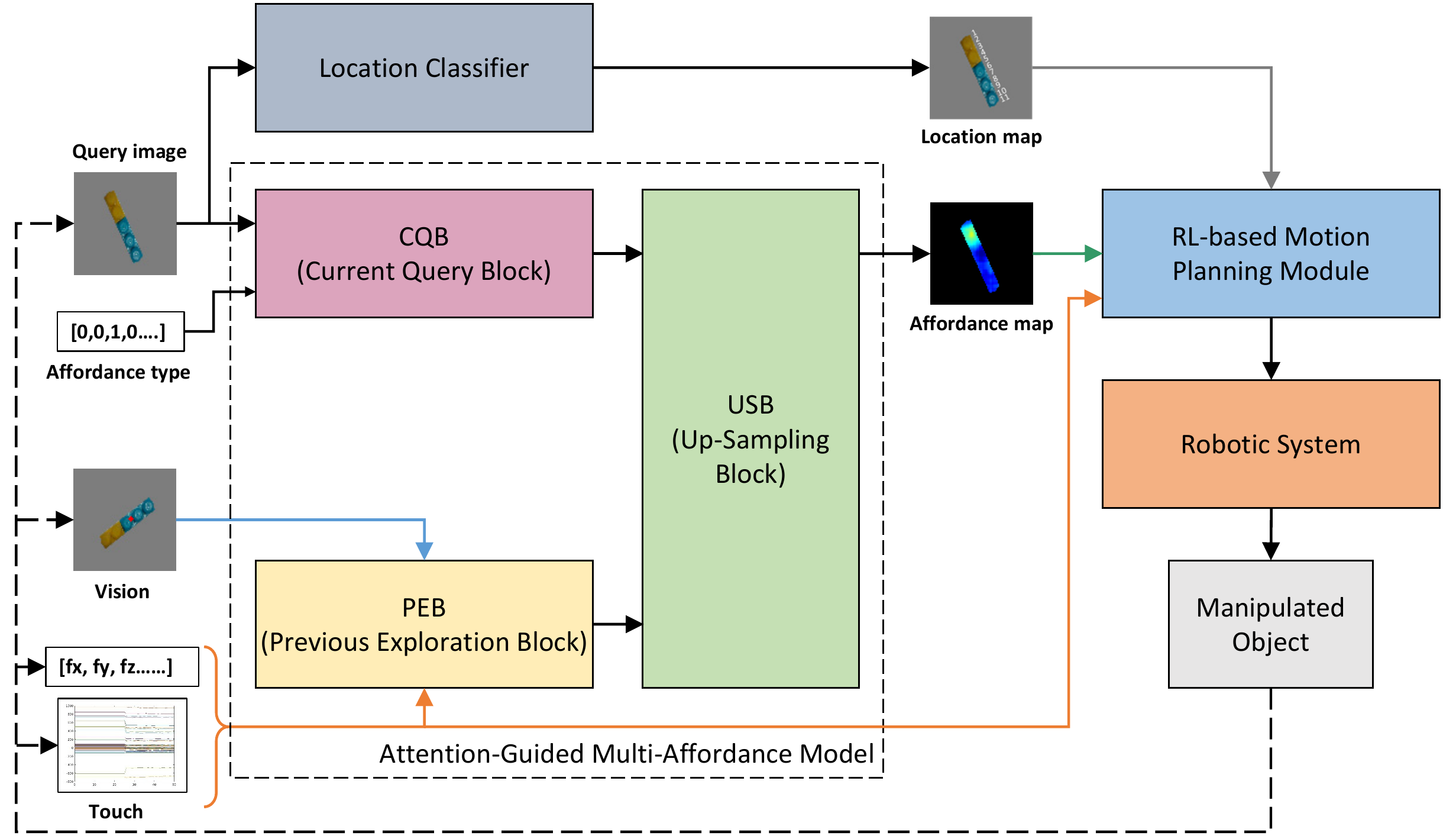}}\vspace{-0.3cm}
\caption{Overall object manipulation planning framework.}
\label{fig:framework}\vspace{-0.55cm}
\end{figure}

The overall object manipulation planning framework is proposed and depicted in Fig. \ref{fig:framework}, which mainly consists of a multi-affordance representation model, a location classifier, and an RL-based motion planning module. The affordance model is designed to predict the potential manipulable locations of an affordance, given an object with unknown configuration and in random orientation. The location classifier is designed to divide the object into several labelled regions based on visual information. This classifier, trained with AlexNet \cite{Krizhevsky2012ImageNetCW}, a shallow CNN on our manually labeled image-position pairs, provides the robot a discrete action space during object manipulation. The RL-based motion planning module is designed to plan the optimal motion such that the robot end-effector can reach the desired manipulable location with minimum steps. Significantly, the affordance model, fusing the vision and multimodal touch information, is used as an input state to the motion planning module. This affordance model predicts and highlights the manipulable information on top of the object image data, which allows faster learning convergence of the motion planning module. Note that with the motion planning module guiding the next explorations, the collected information is more meaningful and valuable than using a random exploration strategy. As the value of exploration increases, it will improve the accuracy of affordance predictions and reduce the number of explorations required for a successful grasp or push. Note that this work focuses on the affordance model and motion planning module, which are detailed in the following.

\subsection{Multisensory Multi-Affordance Representation Model}
%The deep learning process of this model training is an end-to-end supervised learning.
The training process of this deep learning model is an end-to-end supervised learning.

\subsubsection{Problem Formulation}
The objective of this model is to generate an affordances map for manipulation by an end-to-end deep learning (DL) network, given inputs which include a visual image of an object, required affordance type and previous explorations data (manipulation locations and explorations outcome). %Similar to \cite{Matthewaffordanceprediction2020}, objects focused in this work are restricted to be simple and planar. 
The two challenges of our task are that objects of the same appearance and shape may (i) have different mass distributions (different configurations of extra weights); (ii) have different responses for each affordance type, which lead to different manipulable regions for objects with the same visual appearances. To overcome these challenges, physical interaction with the objects is essential. This interaction allows exploration of the intrinsic properties of the object in order to form prior data to the model. The information obtained from an RGB camera, force/torque and tactile sensors is used as inputs to the DL network to predict manipulable regions for an affordance type.

In this work, an affordance map $M$ is represented as pixel-wise probability. The pixel in affordance map associated with the highest value is considered to be the most suitable manipulation location given a query image $I_q$ of an object and affordance type $A_q$.  Each previous exploration consists of an image $I_{exp}$ of the object and the manipulation location $Rs$ containing pixels within a circle with a radius of 5 pixels. The manipulation location is labelled in different colours: red for failure and green for successful manipulations, respectively. Previous explorations on the same object under the same affordance type are represented as $P(I_{exp}^{(1)},I_{exp}^{(2)}...I_{exp}^{(k)}, R_s^{(1)},R_s^{(2)}...R_s^{(k)})$. Besides that, the sensory information of the previous attempts contain forces/torques $FT$, tactile $Ta$ and a Boolean variable $o$ denoting whether the exploration was successful or not. Thus, the $i^{th}$ exploration is represented as $e_i$=$\{P_i, FT_i,Ta_i, o_i\}$.

\subsubsection{Attention-Guided Network Design}
\begin{figure}[!b]
  \centering\vspace{-0.5cm}
  \includegraphics[width=1.0\columnwidth]{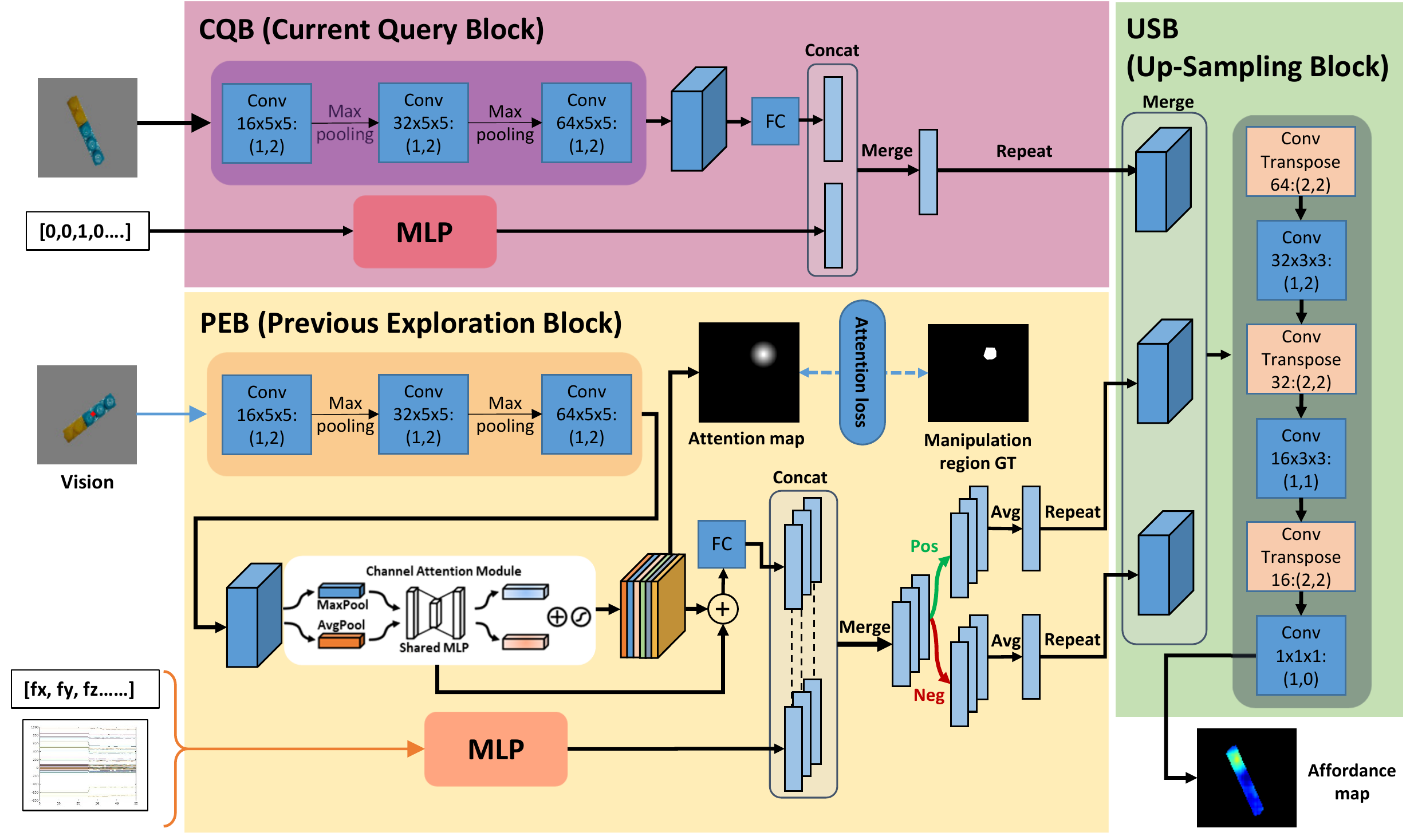}\vspace{-0.3cm}
\caption{Main architecture of AGN.}
\label{fig:main_archi}
\end{figure}

To generate the affordance map $M$, the network is trained to fuse visual information, multimodal touch information, affordance type of the current query, and the interaction experience of previous explorations of the object.
%Overall, our work is the extension of the work presented in \cite{Matthewaffordanceprediction2020} that solves the problem of robotic grasping with sensory feedback.
Fundamentally, the proposed network draws inspiration from \cite{Matthewaffordanceprediction2020} and is akin to a semantic segmentation network \cite{FCN2015,MASKRCNN2017} which predicts a class for every pixel in an input image. We focus only on single manipulation points in our task. A manipulation location in an exploration/attempt is either positive (successful support exploration) or negative (failure support exploration). The proposed DL network in this paper aims to extract meaningful context from support explorations, so that it is able to predict an affordance map given a new image of the same object. 
%of two cells with extra mass is in low success rate, especially for non-symmetric objects. (this is due to prediction model prefers choosing successful manipulation point near the object's geometric center \cite{Matthewaffordanceprediction2020}; once the real mass center of an object deviate its center a lot, it will be challenging for prediction model to make correct manipulation point prediction.)
To overcome the aforementioned challenges and further improve the affordance prediction performance, the proposed DL network includes a branch to fuse affordance type information and one channel attention block in previous explorations feature extraction block to allow the network to pay more attentions on the manipulation region. Hence, the proposed DL network is named as an attention-guided network (AGN). The main architecture of AGN is depicted in Fig. \ref{fig:main_archi}. The description of AGN is detailed below. %Subsection \ref{sec:modeldescription}.

%\subsubsection{AGN Description}
%\label{sec:modeldescription}
For the inputs, all the input images are resized to 88$\times$88; the touch sensory information is standardized to have zero mean and unit variance; the affordance type is represented by an one-hot vector, namely, affordance vector. If the number of affordances is $n$ and the indicated affordance type is $l^{th}$ in the predefined affordance list given the current query, the affordance vector will be represented as a vector where all items are 0, except 
that the $l^{th}$ entry is 1. As shown in Fig. \ref{fig:main_archi}, all inputs go through the AGN and output an affordance map visualized by applying $COLORMAP\_JET$ wherefrom red to blue correspond to probability from high to low.

\noindent {{\textbf{Current Query Block (CQB):}}} Given a query image of an object and the requested affordance type, the CQB, which is a combination of a convolutional neural network (CNN) and a multilayer perceptron (MLP), is used to extract and fuse visual features and affordance query information. The CNN contains three convolutional layers, each of them has a filter with a kernel size of $5\times5$, a stride of 1, a padding of 2 and followed by a max pooling layer. At last, there is a \emph{fc} (fully connected) layer to convert the feature map to a feature vector. The output of CNN is a feature vector with a size of 64$\times$1. The MLP contains a hidden layer which linearly converts the dimension of the vector from $n$ (number of affordance type) to 64. The outputs of CNN and MLP are concatenated and merged to a new feature vector with the size of 64. The vector is then repeated 11$\times$11 times along `height' and `width' dimensions to form a feature map of dimension 64$\times$11$\times$11. 

\noindent {\textbf{Previous Explorations Block (PEB):}} This block includes a CNN, a MLP and a channel attention module. The configurations of the convolutional layers and max pooling layers are the same as those of the CNN designed in CQB. The only difference is that there is no \emph{fc} layer after the CNN, and thus the output of CNN is a 64$\times$11$\times$11 feature map. This feature map is fed into the channel attention module that is optimized by minimizing attention loss between attention map and ground-truth manipulable region. The reason for designing an attention module here is that the images of previous explorations are of different orientations, poses, and manipulable locations. However, the difference of manipulable locations on an object is the most important factor to affect the exploration result. Hence, special attention on the locations' difference is helpful for the AGN to learn the best manipulation location on the current image, from experiences where different manipulable locations result in different exploration results. The feature map after the attention module is fed into a \emph{fc} layer to generate a 64-dimensional vector. The MLP also contains a hidden layer that converts the vector dimension from the number of sensory information to 64. For each previous exploration, the feature vectors from vision and touch information branches will be then concatenated and merged into a new feature vector with a dimension of 64. The new vectors of all previous explorations will be split into two groups, positive and negative, corresponding to the exploration results (i.e., success and failure). Subsequently, the average representation for each group, producing one positive and another negative representative vectors, is computed. These vectors are then repeated 11$\times$11 times to form the feature maps with a dimension of 64$\times$11$\times$11.   

\noindent {\textbf{Up-Sampling Block (USB):}} The outputs of CQB and PEB are three feature maps, which are then concatenated and bi-linear interpolated \cite{SMITH1981201} along concatenated dimension to finalize features for subsequent USB. The USB consists of three groups of ConvTranspose and Conv2d layers. The ConvTranspose layer with a stride of 2, and a padding of 2 is an up-sampling layer which does not change the channel units but doubles the feature map size. The first two Conv2d layers are with filters (whose kernel size is 3, stride is 1, and padding is 1) and they halve the channel unit and keep the feature map size constant. The last Conv2d layer has a filter with a kernel size of 1, a stride of 1, and a padding of 0, followed by a sigmoid layer to output a 2D affordance map with a dimension of 1$\times$88$\times$88.

\subsubsection{AGN Optimization}
 The objective function is composed of a binary cross-entropy loss (on current frame single manipulation point)  and $k$ attention losses (on prior exploration manipulation areas) which are mean squared error (MSE) losses as is defined below:
 \begin{equation}\label{eq:loss_function}
\textsl{L}=\mathit{t} log(M)+(1-\mathit{t}) log(1-M)+\lambda\frac{1}{N}\sum_{i=1}^{k}\sum_{j=1}^{N}(y_{ij} - {\hat{y}}_{ij})^2,\hspace{-0.1cm}
\end{equation}
where $t$ is the ground truth map of single manipulation points with pixel value of either 0 (failed manipulation) or 1 (successful manipulation), while remaining points are with values other than 1 or 0 . $M$ is the predicted affordance map, $N$ is the number of pixels in each map, $k$ is the number of prior explorations, while $y_{ij}$ and ${\hat{y}}_{ij}$ are observed and predicted value on the $j^{th}$ pixel of $i^{th}$ prior exploration, respectively. $\lambda$ is the loss balancing parameter.

The AGN weights are optimized using Adam optimizer \cite{Adamoptimizer2015} in the training process, and the learning rate is set as $5e^{-4}$ while the weight decay is $2e^{-4}$. The model training epoch is 1000. Data augmentations are implemented for training on the visual inputs for both CQB and PEB, such as horizontal and vertical flips. 

\subsection{Deep Reinforcement Learning-based Motion Planning}
The motion planning problem in this paper can be described as a  Partially Observable Markov Decision Process (POMDP) \cite{BrechtelITS2014}. The process is defined by the functions of ($S$, $A$, $\mathcal{T}$, $R$, $O$), where $S$ is the collection of environment states which is the combination of predicted affordance map from the AGN model and location map from the location classifier in our method, $A$ is the set of allowed actions given a state, $\mathcal{T}$ is the transition function from the current state ($s_t$) to the next state ($s_{t+1}$) with action ($a_t$) at the current state, $R$ is the scalar reward computation given a state-action pair ($s$, $a$), $O$ is the tactile sensor feedback computation function. In this work, the tactile sensor input is computed as the temporal difference of three successive tactile information. The problem in the POMDP is that the current actions affect the next states and the future rewards, this can be solved via using DRL algorithms. In this work, the value-based algorithm, namely, deep Q-learning (DQN) is employed as the DRL algorithm to train an agent, the Q-value update policy with learning rate $\gamma$ is expressed in 
\begin{equation}\label{eq:qlearning}
    Q(s,a)=Q(s,a)+\alpha[r+\gamma \underset{{a}'}{\max}Q({s}',{a}')-Q(s,a)].
\end{equation}

Given the current affordance map and touch information, %the agent is to predict moving direction and step size to localize the successful manipulation points within shortest steps. 
the agent training process is designed like in works \cite{XuICRA2021,fangicip2021} to make optimal prediction on two actions,  moving direction  with action space of 8 (\textit{left, right, up, down, left-up, left-down, right-up, and right-down}) and moving step size with action space of 5 simultaneously. Thus, the output Q of the DQN contains three branches:
\begin{align}\label{eq:qlearning}
   Q(s,a; \theta) = &  Q^s(s;\theta^f,\theta^s)+Q^{md}(s,a^{md};\theta^f,\theta^{md}) \nonumber \\
   & +Q^{ms}(s,a^{ms};\theta^f,\theta^{ms}),
\end{align} 
where $Q^s$ is the state value, $Q^{md}$ is the moving direction Q-value, $Q^{ms}$ is the moving step size Q-value. The three branches share the same feature extractor {(similar to the one used in the CQB and PEB of AGN)} and inputs, and the $\theta^f$ is the parameters of the feature extractor, $\theta^s$, $\theta^{md}$, and $\theta^{ms}$ are parameters of the three branches. $\theta$ is a collection of these parameters where $\theta=\lbrace \theta^f,~ \theta^f,~\theta^{md},~\theta^{ms}\rbrace$. 

The agent is optimized in the training process by minimizing the difference between the so-called current Q-value and target Q-value computed from each ($s$, $a$) pair. To optimize the agent's performance, the following RL loss is applied, 
\begin{align}\label{eq:TD_error}
    L(\theta)= &\underset{\textrm{target-Q}}{\underbrace{[r(s_t,a_t)+\gamma \underset{a_{t+1}}{\max}Q(s_{t+1},a_{t+1};\theta)}}-Q(s_{t},a_t;\theta)]^2.
\end{align}

Once an action is determined at a certain state, the reward after moving $p$ steps can be determined by
\begin{equation}\label{eq:reward_function}
    r(s,a)=\begin{cases}
 1+\frac{1}{1+e^p}, ~$if success manipulation location found$\\ 
 0, ~\qquad \quad $otherwise.$
\end{cases}
\end{equation}
After training, the parameters are optimized, the approximate action decision given current state can be obtained by
\begin{equation}\label{eq:action_policy}
 %   \left\{\begin{matrix}
a_{t}=\arg\underset{a\in A_{t}}{\max}Q(s_{t-1},a,\theta). %\\ 
%a_{t}^{r}=arg\underset{a^r\in A_{t}^{r}}{max}Q_r(s_{t-1},a^r,\theta)
%\end{matrix}\right.
\end{equation}
The output of the DRL framework is the action decision according to affordance map of an object and sensory feedback.% after the agent taking the action decisions, an target location is generated for robot manipulation.

\section{Experiments and Results}
To verify and validate the effectiveness of the proposed framework, several experiments are conducted using the dataset published in \cite{Matthewaffordanceprediction2020} and our dataset collected on the robotic system shown in Fig. \ref{fig:robot}. A use case on object manipulation is also demonstrated with the robotic system.

\subsection{Experimental Setup}
For the evaluation of the proposed affordance model, we benchmark the performance of the proposed AGN against the model presented in \cite{Matthewaffordanceprediction2020} (namely, baseline model) on both the YCBUSR dataset \cite{Matthewaffordanceprediction2020} and our dataset. Remarkably, our dataset is collected on a robot with a 2-finger gripper instead of a suction cup. Unlike the suction cup that is able to make contact with a large surface of the object, grippers can only manipulate certain parts of the object. This results in the collected samples for each object being sparse in the 2-finger gripper scenario. In such case, the average grasp successful rate of our grasping setup is 17.37\%, which is much lower than that of the YCBUSR dataset, 47.13\%. Hence, our dataset provides a more challenging scenario for affordance learning. 

On both datasets, both the baseline and our models are evaluated based on two settings: (i) how well a model can predict affordances if it has seen an object's appearance before but has no knowledge about a particular mass distribution; and (ii) how well a model can generalize knowledge from seen objects to unknown objects (`seen' and `unseen', respectively hereafter). All evaluations use five-fold cross-validation. As depicted in \cite{Matthewaffordanceprediction2020}, the optimal affordance prediction results considering both accuracy and consumed time can be achieved when there are five previous explorations. Therefore, in the experiments on the affordance model, the number of previous explorations for model benchmarking is manily set as 5. Further on, the overall framework is evaluated on our dataset with comparison of the RL-based motion planning module with different state inputs. The training and evaluation of AGN is under PyTorch \cite{paszke2017automatic} framework on a machine with one NVIDIA GeForce GTX 1080 GPU.

\subsection{Experiments on Affordance Representation Model}
\subsubsection{Results on YCBUSR dataset}
The YCBUSR dataset contains nine classes of objects, with a total of 2868 samples. The same rules used in \cite{Matthewaffordanceprediction2020} are followed to split objects and classes for 5-fold validation. The benchmark results on the YCBUSR dataset are evaluated by the mean and standard deviation (SD) of the Area Under the Receiver-Operating Characteristic Curve (AUROC) across all folds of cross-validation. %, which is the same as the metrics used in \cite{Matthewaffordanceprediction2020}. 
The results are shown in Table \ref{tab:evaluation results on YCBUSR}. From the table, it is clear that our AGN obtains higher AUROC than the baseline model in both `seen' and `unseen' settings with different numbers of explorations.  

\begin{table}[!b]
\renewcommand{\arraystretch}{1.1}\vspace{-0.5cm}
\caption{Benchmark results (in AUROC: mean$\pm$sd) on YCBUSR dataset}\vspace{-0.3cm}
\centering
\label{tab:evaluation results on YCBUSR}
\begin{tabular}{>{\centering\arraybackslash}p{1.8cm}|>{\centering\arraybackslash}p{1.2cm}>{\centering\arraybackslash}p{1.2cm}>{\centering\arraybackslash}p{1.2cm}>{\centering\arraybackslash}p{1.2cm}}
\hline
                         & \multicolumn{2}{c}{baseline \cite{Matthewaffordanceprediction2020}} &\multicolumn{2}{c}{AGN}   \\ \hline
\begin{tabular}{@{}c@{}}\# of explorations \end{tabular}& 4& 5 & 4 & 5\\           \hline             
seen   & 0.903$\pm$0.03 & 0.910$\pm$0.03     & 0.913$\pm$0.03     & 0.938$\pm$0.03          \\ \hline
unseen & 0.898$\pm$0.02 & 0.905$\pm$0.02     & 0.907$\pm$0.03     & 0.930$\pm$0.03          \\ \hline
\end{tabular}
%\vspace{-0.5cm}
%\end{adjustbox}
\end{table}

\begin{figure}[!t]
  \centering\vspace{0.2cm}
  \centerline{\includegraphics[width=1.0\linewidth]{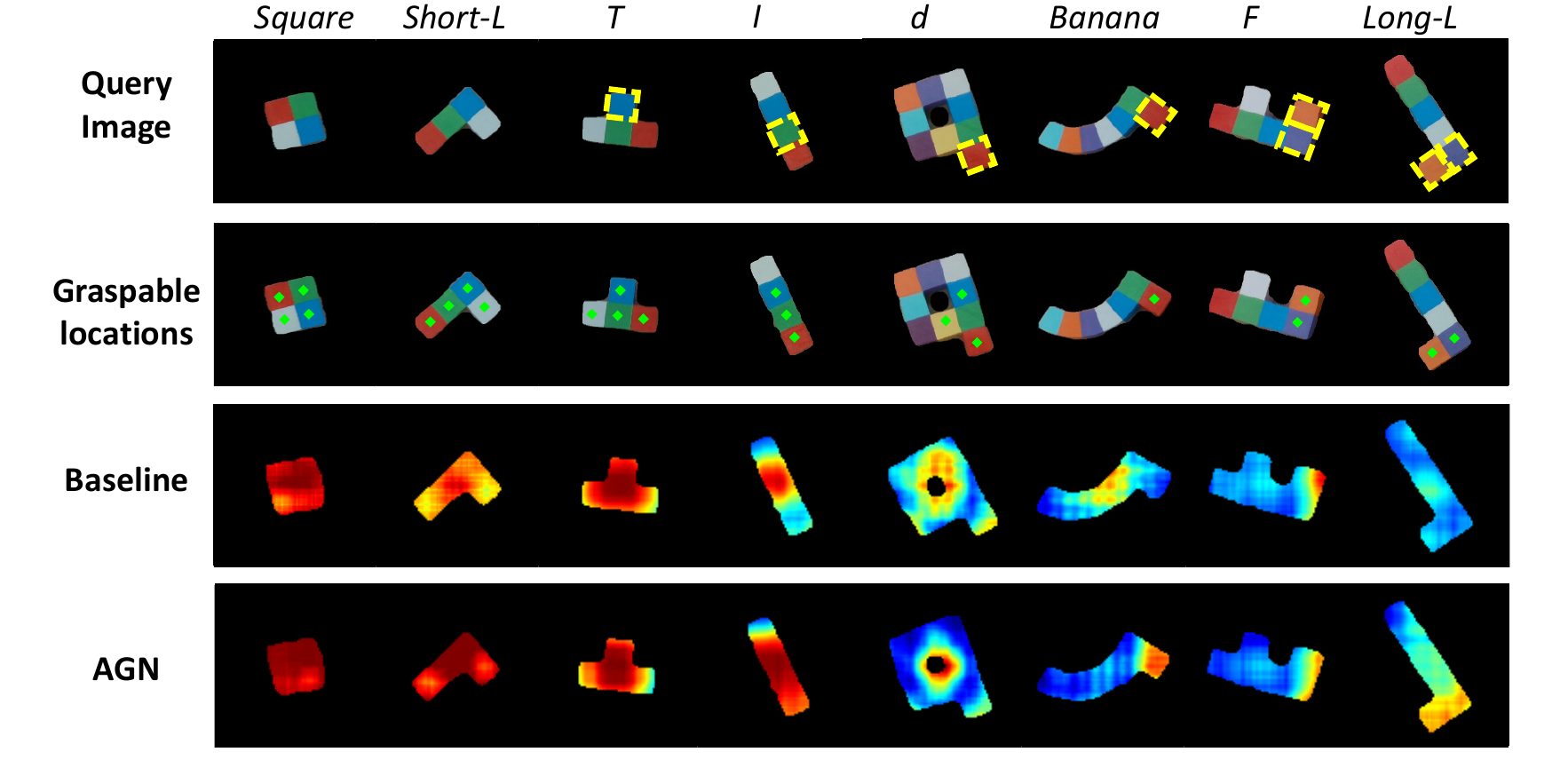}}\vspace{-0.3cm}
\caption{Sample affordances predicted with five previous explorations: extra weights are located in the dotted yellow boxes; green dots are (successful) graspable locations on the objects.}\vspace{-0.5cm}
\label{fig:YCBUSR_samples}
\end{figure}

Sample affordances predictions on the YCBUSR dataset can be found in Fig. \ref{fig:YCBUSR_samples}. The masses of $Square$ and $Short-L$ shape objects are uniform and all previous explorations are successful grasp. Thus, almost all pixels on the affordance maps predicted by both baseline and AGN are in red or yellow (relative high probability). Both the $T$ and $I$, as well as $d$, and $Banana$ shape objects contain one cell with an extra weight. For the $T$ shape, the extra weight is on its symmetry axis, therefore both models are able to generate sensible affordance maps easily. However, for the $I$, $d$, and $Banana$ shapes, the cell with extra weight is biased to one side of the object. Based on the previous exploration experiences, the affordance map generated from our model is more convincing. The $F$ and $Long-L$ shape objects in the last block each contain two cells with extra weights biased to one end of the object, which make their mass distribution non-uniform and non-symmetric. Affordance map generation under this situation is very challenging, and it needs the model to pay more attentions to the previous explorations. Both results shown in the table and figure clearly show that our AGN model outperforms the baseline model, which indicates that the attention module in the PEB improves performance. It is also noteworthy that the accuracy on both `seen' and `unseen' objects using the AGN with four explorations are equivalent to those using the baseline model with five explorations.  In other words, the proposed AGN requires fewer explorations (20\% less in this case) than the baseline model to achieve a competitive high accuracy (e.g., > 0.9). This implies that the proposed AGN has the potential to improve the efficiency.

%From the prediction results of both baseline and our model, we can clearly see that the AGN outperforms. This is attributed to the attention module in the PEB block of the AGN.

\subsubsection{Results on Our Dataset}
\begin{figure}[!b]
    \centering\vspace{-0.6cm}
    (a) \subfigure{\includegraphics[width=0.6\columnwidth]{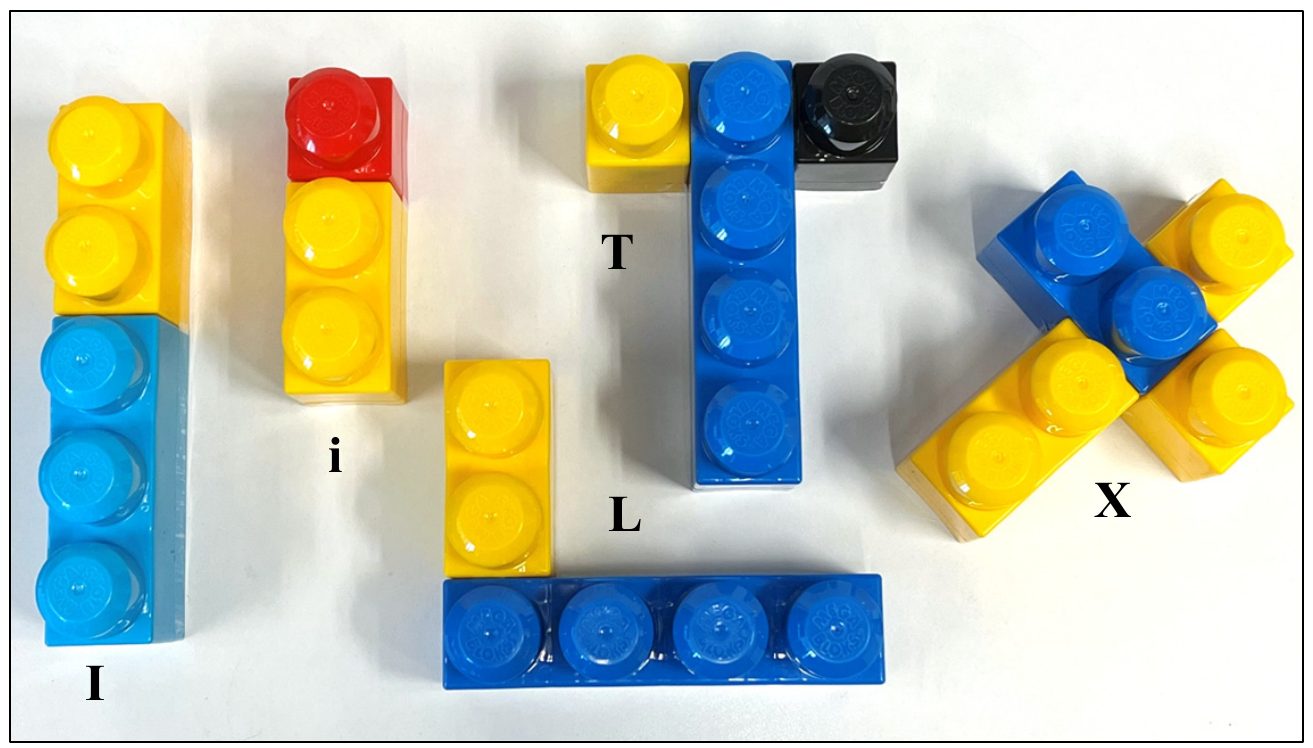}}\\
    (b) \subfigure{\includegraphics[width=0.6\columnwidth]{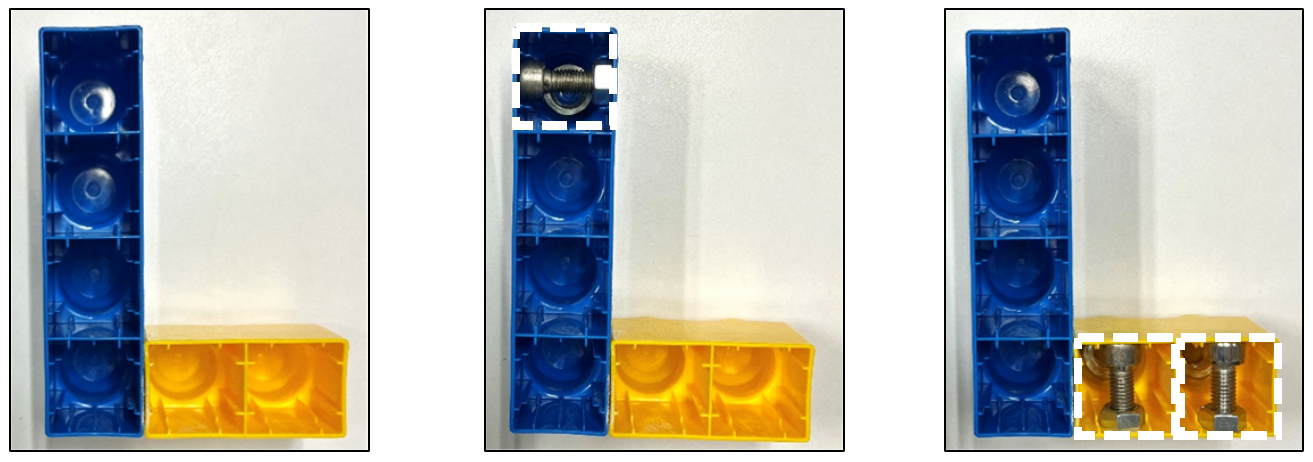}}\vspace{-0.3cm}
    \caption{Objects used in this work: (a) all object shapes; (b) different configurations of the \textbf{L} shape object with white dash-line boxes indicating locations of extra weights.}
    \label{fig:object}
\end{figure}

% \underline{System Setup:}
%\textcolor{red}{For convenience of comparison with %manipulable location prediction from the overall framwork, in our dataset, both the AUROC and success rate (SR) are used as evaluation metrics on all methods.} 
The robotic system shown in Fig. \ref{fig:robot} is used for both data collection and object manipulation in the following experiments. Five objects with different shapes constructed by the combinations of different building blocks are used to represent the real-world objects (as shown in Fig. \ref{fig:object}). They are labeled as \textbf{I} (long stick), \textbf{i} (short stick), \textbf{T},  \textbf{L}, and \textbf{X} (cross). Also, each object has three different configurations of mass distributions by adding the hidden weights into the hollows of the blocks. Fig. \ref{fig:object}(b) shows example configurations of the \textbf{L} object. In total, 2442 and 3234 samples are used in grasping and pushing, respectively.\vspace{0.1cm}

\noindent\underline{\emph{Experimental Results on Object Grasping}}\\
Evaluation results on our dataset are first benchmarked between the baseline model and AGN based on `seen' and `unseen' setting using the same AUROC metrics on grasp affordance. We evaluated the performance of our model for force/torque only, and force/torque with tactile inputs. %The proposed AGN trained only with tactile data is denoted as AGN-T, while 
The proposed AGN trained with additional 48-dimensional tactile data on top of the 6-dimensional force/torque data is denoted as AGN-T.

\begin{table}[!t]
\centering\vspace{0.2cm}
\renewcommand{\arraystretch}{1.1}
\caption{Benchmark results (in AUROC: mean$\pm$sd) on our grasping dataset}\vspace{-0.3cm}
%Grasping affordance prediction benchmarking results (in AUROC) on ourdataset
\label{tab:grasping evaluation results on ourdataset}
\begin{tabular}{c|>{\centering\arraybackslash}p{2cm}>{\centering\arraybackslash}p{2cm}>{\centering\arraybackslash}p{2cm}}
\hline                              
& baseline \cite{Matthewaffordanceprediction2020} & AGN       & AGN-T    \\ \hline
seen    & 0.743$\pm$0.03     & 0.768$\pm$0.03         & 0.792$\pm$0.03     \\ \hline
unseen  & 0.721$\pm$0.04     & 0.754$\pm$0.03          & 0.769$\pm$0.03\\
%\hline 
%\multirow{2}{*}{\begin{tabular}{@{}c@{}}Challenge\\configs\end{tabular}} & seen    & 0.$\pm$0.     & %0.$\pm$0.          & 0.$\pm$0.    \\ \cline{2-5} 
%&unseen  & 0.$\pm$0.     & 0.$\pm$0.          & 0.$\pm$0. \\
\hline
\end{tabular}\vspace{-0.3cm}
\end{table}

\begin{figure}[!t]
  \centering
  \centerline{\includegraphics[width=1.0\linewidth]{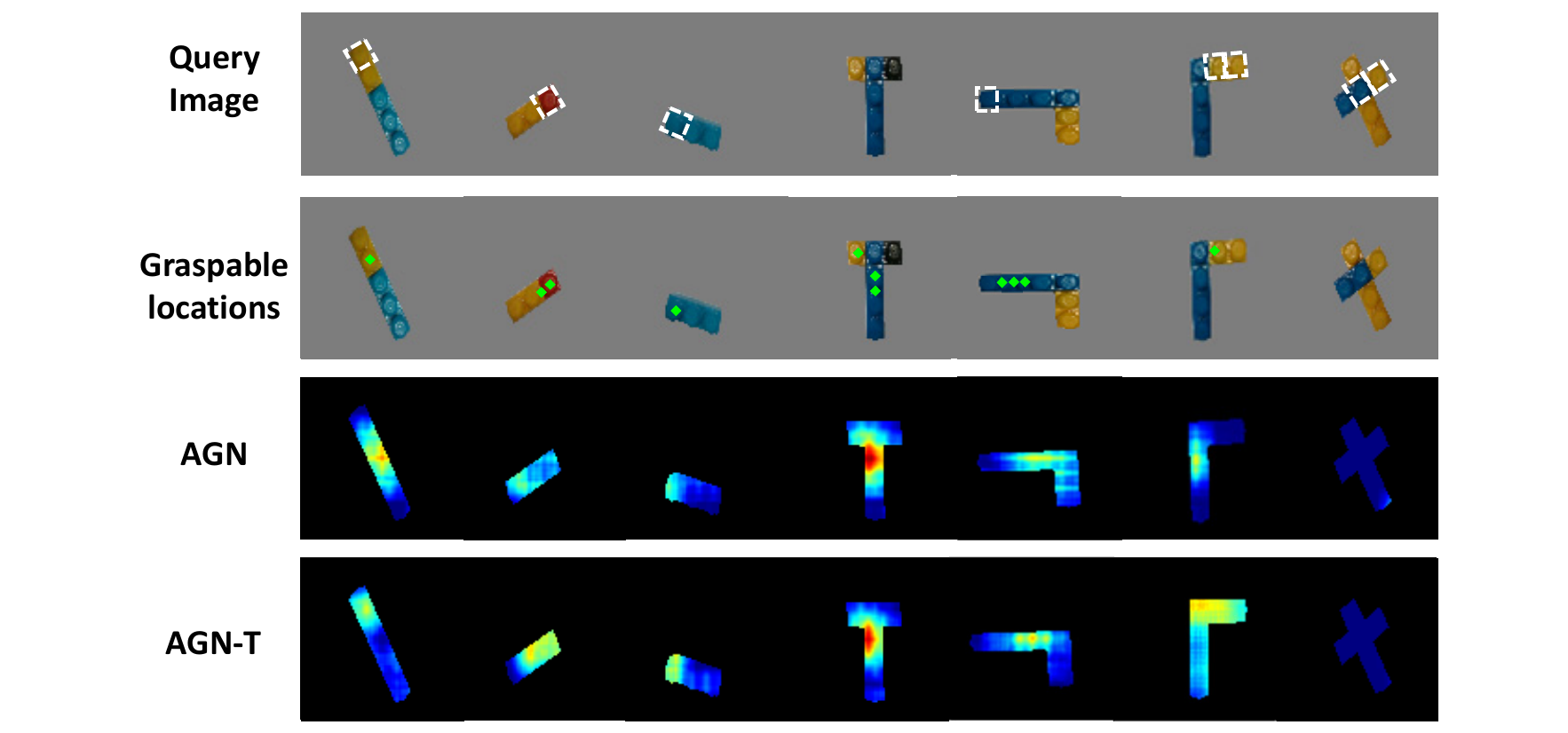}}\vspace{-0.3cm}
\caption{Sample grasp affordance maps generated on our dataset: extra weights are located in the white dash-line boxes.}\vspace{-0.6cm}
\label{fig:ourdataset_samples_grasping}
\end{figure}

The results are tabulated in Table \ref{tab:grasping evaluation results on ourdataset}. As listed in the table, the AUROC on our dataset for both the baseline and our models are lower than that on the YCBUSR dataset. This is reasonable because our dataset is more challenging than the YCBUSR dataset as mentioned previously. Nevertheless, consistent improvement in results can be observed for both using our model against baseline and using additional tactile inputs against only single-point force/torque data. Moreover, it is evident that the tactile sensing information can definitely improve the model prediction accuracy in comparison to those without the tactile data. Furthermore, some tests are carried out using the proposed AGN trained only with tactile data (no force/torque data), which show results of $0.803\pm0.04$ and $0.756\pm0.04$ (in AUROC) on `seen' and `unseen' objects, respectively. These imply that tactile data provides more information about the affordance than force/torque data. 

%To better understand the effectiveness of the proposed AGN, the results in AUROC on the challenging configurations where the objects' CoMs are far from their centroids are further summarized as listed in Table \ref{tab:grasping evaluation results on ourdataset}. It is obvious that the AGN has a significant improvement in comparison to the baseline model. This again proves that the attention mechanism can help the system to achieve better inference accuracy and robustness.

Sample grasp affordance predictions on our dataset can be found in Fig. \ref{fig:ourdataset_samples_grasping}. It is evident that AGN-T makes more accurate predictions on the appropriate areas for stable grasping. It is also worth noting that the \textbf{X} shape object with the configuration shown cannot be stably grasped at all (i.e., not graspable) due to the limitation of gripper aperture. Interestingly, AGN-T gives a very accurate prediction in this case. This helps inform the robot to plan for other actions in place of graspability to successfully manipulate the object. \vspace{0.1cm} %In summary, the proposed model (especially, with tactile sensing information) achieves better grasping affordance prediction.

\noindent\underline{\emph{Experimental Results on Object Pushing}}\\
There are three lower-level affordances associated with the pushing action. Benchmark results of the push affordance prediction on our dataset are shown in Table \ref{tab:pushing evaluation results on ourdataset}. It can be clearly observed that AGN-T performs much better than AGN. This can be mainly attributed to the fact that tactile inputs can provide not only more information about the contact force than single-point force/torque inputs but also extra information about contact configuration (e.g. friction distribution in this case, stress distribution, contact surface area). The tactile sensor is much better at gathering distributed contact information than the force/torque sensor and thus including the tactile sensing information into the AGN can greatly improve the accuracy and robustness on the push affordance prediction. 

\begin{table}[!t]
\centering\vspace{0.2cm}
\renewcommand{\arraystretch}{1.1}
\caption{Benchmark results on our pushing dataset (in AUROC: mean$\pm$sd)}\vspace{-0.3cm}
%Pushing affordance prediction benchmarking results (in AUROC)
\label{tab:pushing evaluation results on ourdataset}
\begin{tabular}{cc|>{\centering\arraybackslash}p{2.1cm}>{\centering\arraybackslash}p{2.1cm}}
\hline
\multicolumn{2}{c|}{}           & AGN & AGN-T \\ \hline
\multicolumn{1}{c|}{\multirow{2}{*}{Translation}}    & seen & 0.448$\pm$0.03 &  0.863$\pm$0.03                    \\ \cline{2-4} 
\multicolumn{1}{c|}{}  & unseen   & 0.431$\pm$0.04 & 0.846$\pm$0.03                  \\ \hline
\multicolumn{1}{c|}{\multirow{2}{*}{Rotate\_CW}} & seen & 0.512 $\pm$0.04& 0.845$\pm$0.03                     \\ \cline{2-4} 
\multicolumn{1}{c|}{} & unseen   & 0.497$\pm$0.03 & 0.812$\pm$0.04              \\ \hline
\multicolumn{1}{c|}{\multirow{2}{*}{Rotate\_CCW}} & seen & 0.491$\pm$0.03 & 0.843$\pm$0.04                     \\ \cline{2-4} 
\multicolumn{1}{c|}{} & unseen   & 0.462$\pm$0.03 & 0.891$\pm$0.04              \\ \hline
\end{tabular}\vspace{-0.2cm}
\end{table}

\begin{figure}[!t]
  \centering 
  \centerline{\includegraphics[width=1.0\linewidth]{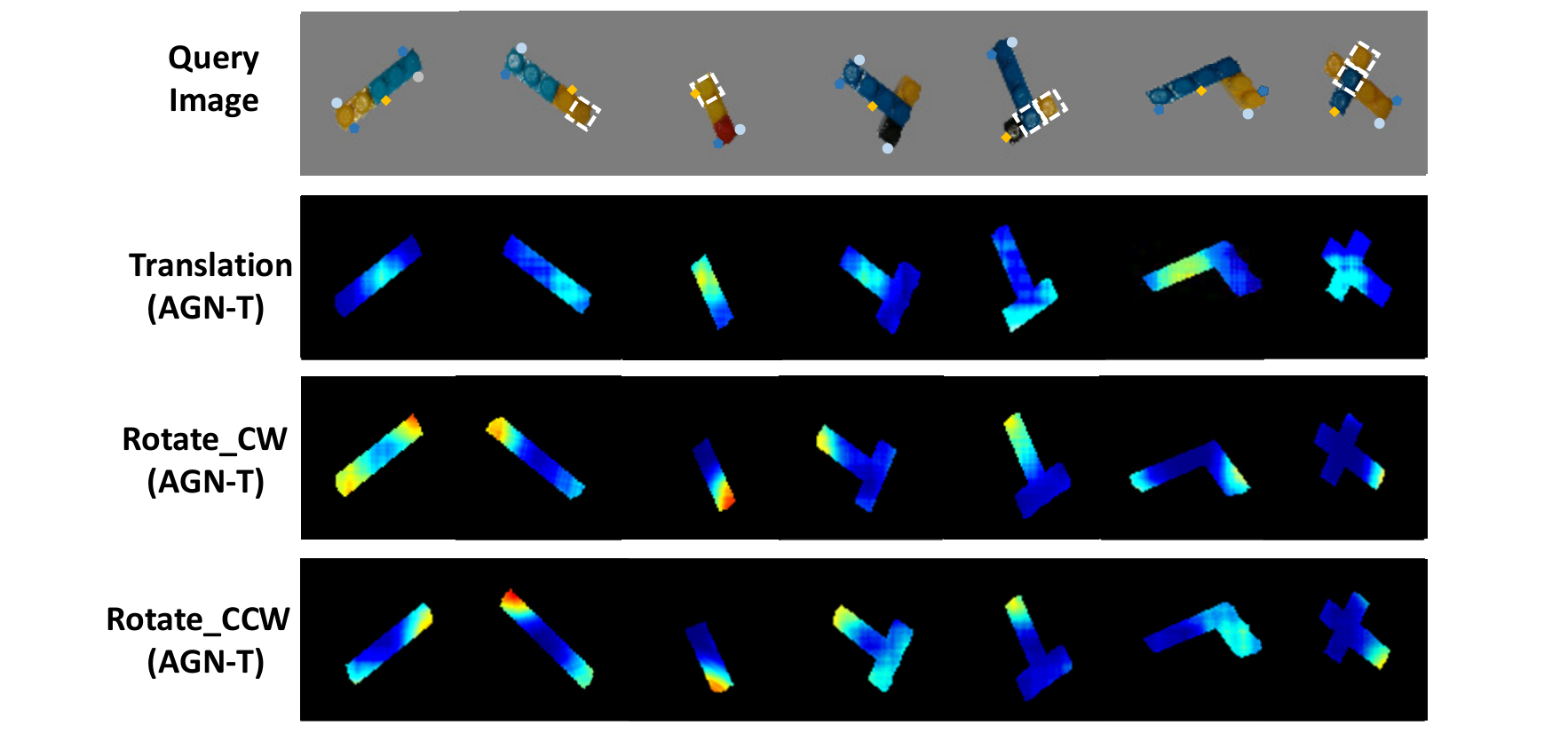}}\vspace{-0.3cm}
\caption{Sample affordances predicted on our dataset for pushing: extra weights are located in the white dash-line boxes, yellow dots are pushable locations for translation, blue dots are CW rotatable locations, grey dot are CCW rotatable locations.}
\label{fig:ourdataset_samples_pushing}\vspace{-0.55cm}
\end{figure}

Figure \ref{fig:ourdataset_samples_pushing} shows sample push affordance predictions on our dataset. As can be seen, the proposed model can well predict the corresponding area for each affordance. By combining with previous results, it can be found that the \textbf{X} shape object is pushable but not graspable. The unified multi-affordance representation model can thus, help the robot to replan the task sequences for object manipulation (e.g., moving an object) if the default action for this task is not affordable.

\subsection{Experiments on the Overall Framework}
The proposed overall framework, affordance representation model (i.e., AGN-T) plus RL-based motion planning module, (namely, AGN-T+RL), is applied on our datasets of both object grasping and pushing. For the object pushing, the performance of the overall framework on translational push is evaluated as it is more frequently used in the packing task mentioned in Section I. For comparison purpose, two frameworks with similar RL structures to the proposed framework but different inputs are implemented and tested. These two frameworks are (i) touch-based RL (without the use of AGN), namely, T+RL; (ii) AGN (without the use of tactile sensor outputs, i.e., only the vision and force/torque information is used) plus the RL-based motion planning, namely, AGN+RL.

\begin{figure}[!b]
  \centering\vspace{-0.5cm}
~~~ \subfigure{\includegraphics[trim=0cm 5.5cm 0cm 0.4cm, clip=true,width=0.9\columnwidth]{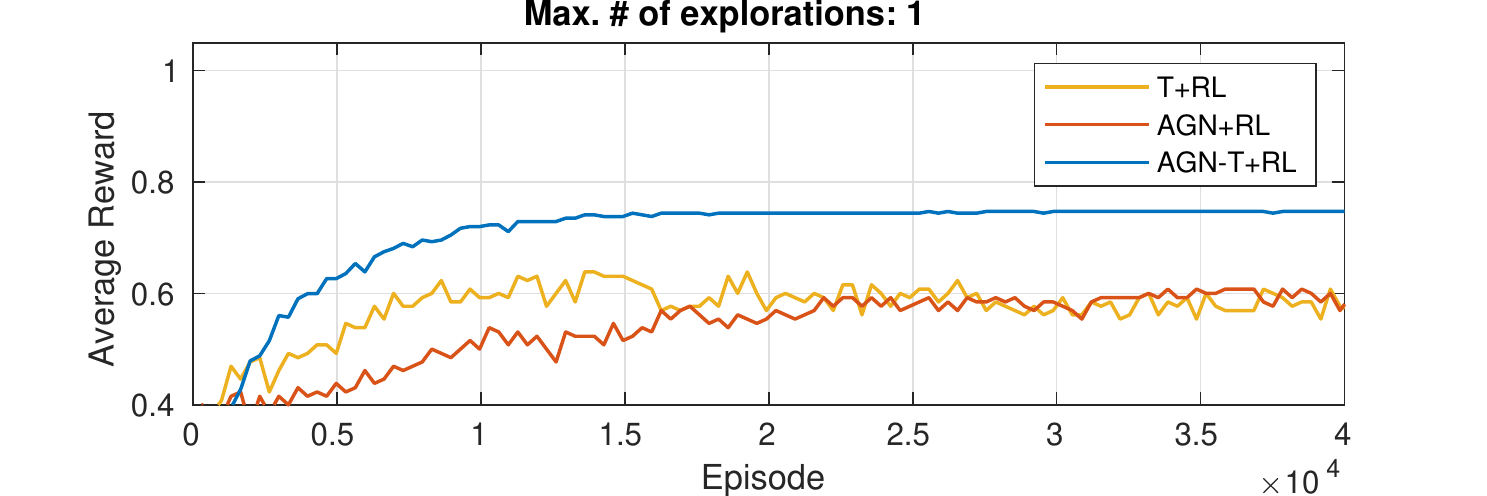}}\\\vspace{-0.1cm}
(a) \subfigure{\includegraphics[trim=0cm 5.5cm 0cm 0.3cm, clip=true,width=0.9\columnwidth]{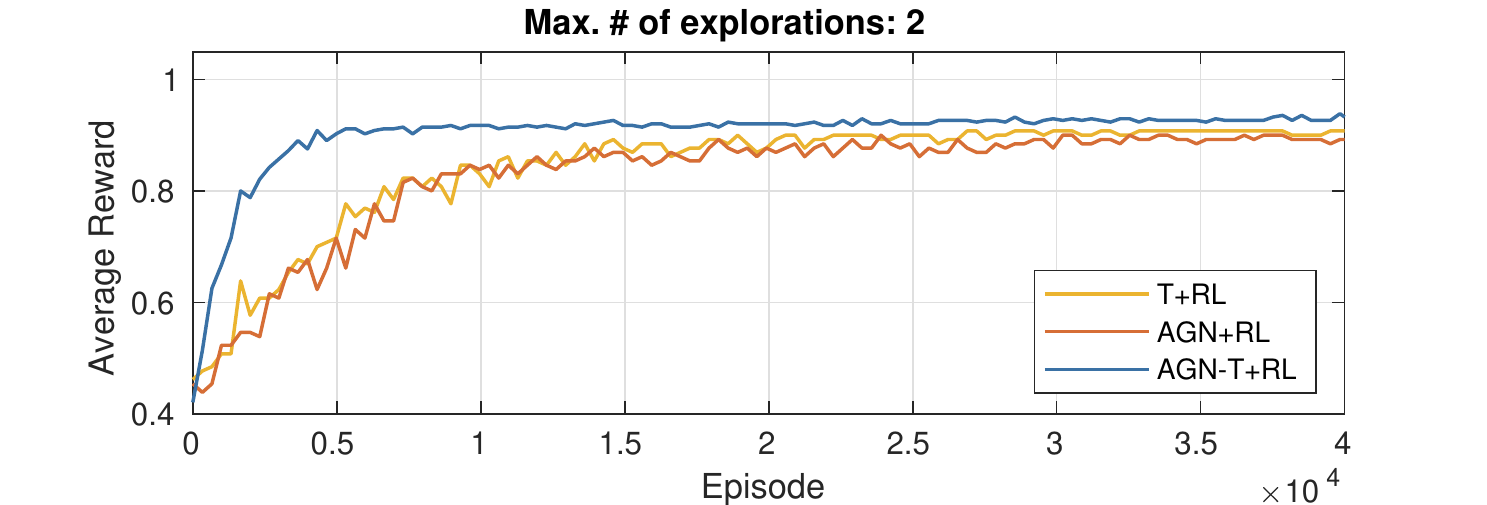}}\\\vspace{-0.1cm}
~~~ \subfigure{\includegraphics[trim=0cm 5.5cm 0cm 0.4cm, clip=true,width=0.9\columnwidth]{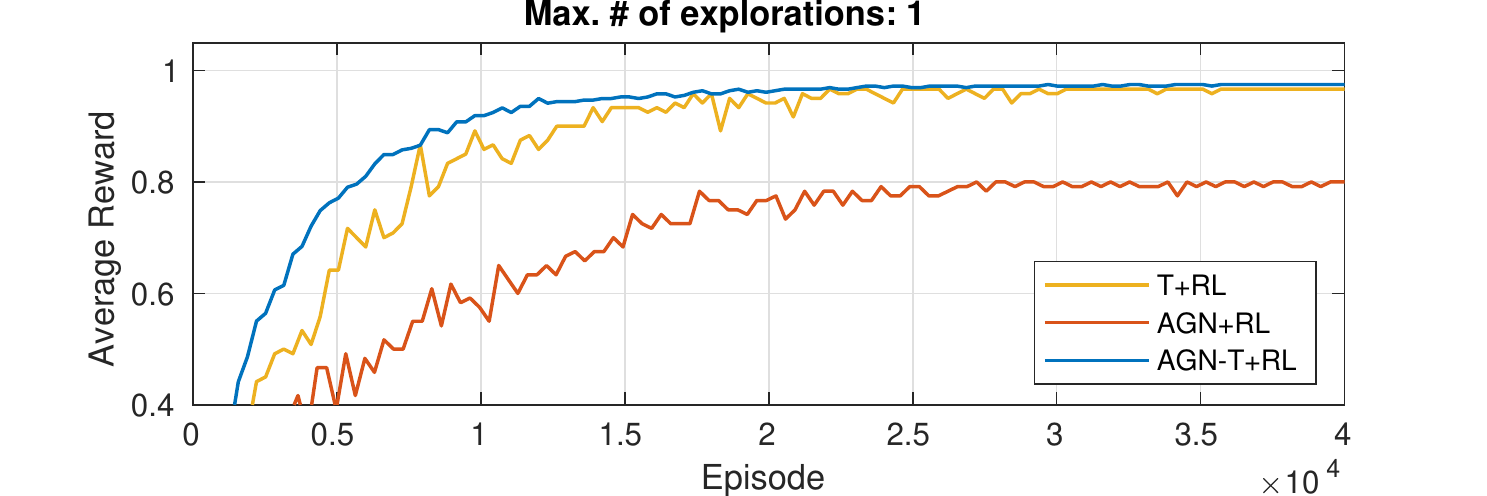}}\\\vspace{-0.1cm}
(b) \subfigure{\includegraphics[trim=0cm 5.5cm 0cm 0.3cm, clip=true,width=0.9\columnwidth]{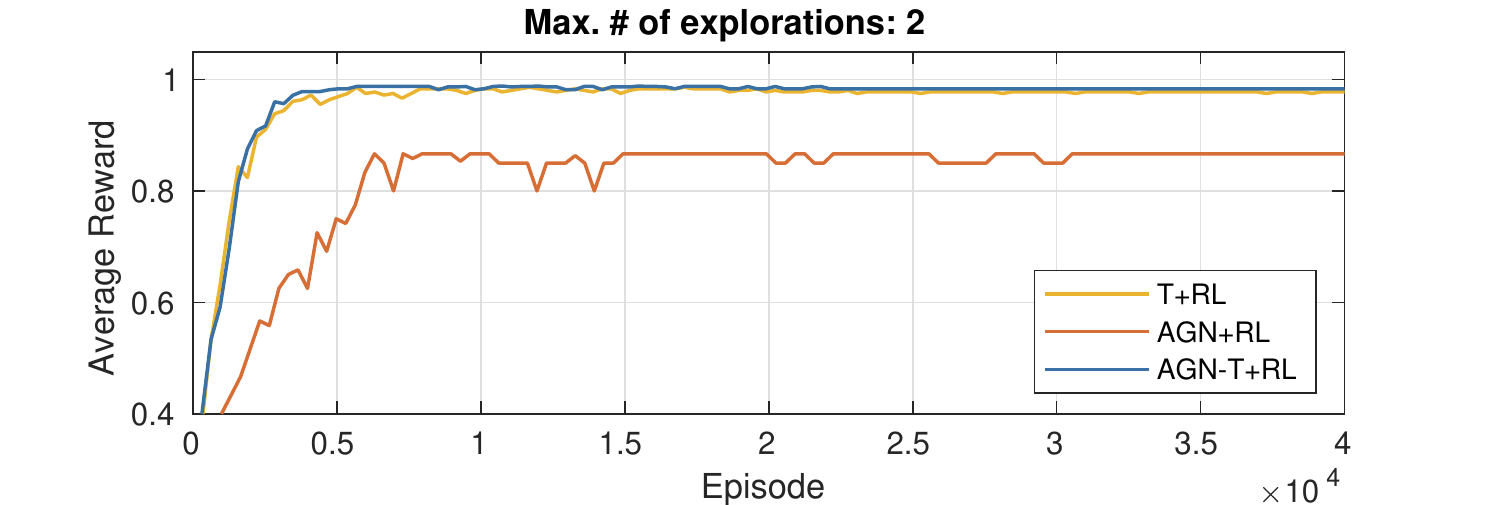}}\\\vspace{-0.2cm}
    \caption{Training process of different RL frameworks on: (a) object grasping (one exploration); (b) object pushing.}\label{fig:reward}%\vspace{-0.3cm}
\end{figure}

The RL training processes of different RL-based frameworks are shown in Fig. \ref{fig:reward}. As can be seen, it is clear that all the RL-based frameworks converge to higher rewards as the number of explorations increases. This also matches the facts found in the affordance model learning, i.e., more explorations benefit to better accuracy. By comparing AGN+RL with AGN-T+RL, it is evident that not only the convergence speeds are slower but also the convergence values are smaller using the AGN+RL, especially on the pushing dataset. This is because the tactile sensor captures far richer contact information comprehensively which is helpful in both predicting the affordance and selecting the correct action. By comparing T+RL with AGN-T+RL, it can be found that the AGN-T+RL converge much faster in the grasping task and slightly faster in the pushing task. It also shows that higher rewards can be obtained using the AGN-T+RL, especially on the grasping dataset with one exploration. These reveal that the generated affordance map by the AGN-T with fusion of the vision and touch information provides useful summarized  information for the system to learn the optimal policy in a shorter time. It should be also noted that the AGN+RL performs comparable convergence speeds and converged rewards to the T+RL on the grasping task. This indicates that the affordance map can compensate for the limited single-point force/torque information (in AGN+RL) to match the rich touch information provided by tactile sensors (in T+RL).

After the RL-models are well trained, the frameworks are then tested on all five objects with different configurations on both grasping and translational pushing. Here, the frameworks without the RL-based module but a simple motion planning module directly utilizing the affordance model (baseline, AGN or AGN-T) outputs are used in the comparison study. The success rates of using different frameworks are listed in Table \ref{tab:success_rate}. It is obvious that the AGN-T+RL achieves the best success rate on both object grasping and translational pushing with two explorations among all the frameworks. By comparing to the frameworks without the use of RL (i.e., baseline, AGN, AGN-T), the AGN-T+RL shows a great improvements on the success rates when they are using the same numbers of explorations (e.g., at least 27\% improvement on grasping and 12\% improvement on pushing for the two-exploration cases). Moreover, it is worth noting that the AGN-T+RL can just use two explorations to achieve higher success rates (at least 6.8\% and 5.9\% improvements on grasping and pushing, respectively) than those frameworks without RL when using five explorations. Thus, it can be concluded that the proposed framework can complete the manipulation tasks with high success rates and high efficiency.

\begin{table}[!t]
\centering
\renewcommand{\arraystretch}{1.1}\vspace{0.2cm}
\caption{Benchmark results on our dataset with motion planning}\vspace{-0.3cm}
\label{tab:success_rate}
\begin{tabular}{>{\centering\arraybackslash}p{1.4cm}|>{\centering\arraybackslash}p{0.6cm}>{\centering\arraybackslash}p{0.6cm}>{\centering\arraybackslash}p{0.6cm}>{\centering\arraybackslash}p{0.6cm}>{\centering\arraybackslash}p{0.85cm}>{\centering\arraybackslash}p{1.05cm}}
\hline
&\multicolumn{6}{c}{Object grasping}\\
\hline
 \# of explorations          & \multirow{2}{*}{\begin{tabular}{@{}c@{}}baseline\\ \cite{Matthewaffordanceprediction2020}\end{tabular}} & \multirow{2}{*}{~AGN} & \multirow{2}{*}{AGN-T} & \multirow{2}{*}{~T+RL} & \multirow{2}{*}{AGN+RL} & \multirow{2}{*}{AGN-T+RL} \\ \hline
1  & 44.48\% & ~46.13\% & ~50.93\% & ~53.57\% & ~~~52.31\% & 60.77\% \\ \hline
2  & 47.52\% & ~51.57\% & ~55.32\% & ~76.96\% & ~~~76.23\% & 83.08\% \\ \hline
5  & 71.94\%  & ~74.37\% & ~76.22\% & - & ~~~- & -  \\\hline\hline
&\multicolumn{6}{c}{Object translational pushing}\\
\hline
1 & - & ~37.58\% & ~72.24\% & ~80.50\% & ~~~62.50\% & 82.50\%\\ \hline
2 & - & ~39.22\% & ~75.63\% & ~85.50\% & ~~~76.04\% & 88.54\%\\ \hline
5 & -& ~43.41\% & ~82.64\% & ~- & ~~~- & - \\\hline
%3 &0.801 & 0.862 \\ \hline
\end{tabular}\vspace{-0.55cm}
\end{table}

% \begin{table}[!t]
% \centering
% \renewcommand{\arraystretch}{1.1}\vspace{0.2cm}
% \caption{Benchmark results on our dataset with motion planning}\vspace{-0.3cm}
% \label{tab:success_rate}
% \begin{tabular}{>{\centering\arraybackslash}p{1.5cm}|>{\centering\arraybackslash}p{1.4cm}|>{\centering\arraybackslash}p{1.2cm}>{\centering\arraybackslash}p{1.2cm}>{\centering\arraybackslash}p{1.2cm}}
% \hline
%                           & \# of explorations & \multirow{2}{*}{Grasping} & \multirow{2}{*}{Pushing} & \multirow{2}{*}{Total} \\ \hline
% \multirow{2}{*}{baseline \cite{Matthewaffordanceprediction2020}}  & 2 & & - & -    \\ \cline{2-5} 
% & 5 & 71.94\%         & -       &  - \\
% \hline
% \multirow{2}{*}{AGN}& 2 & & &\\ \cline{2-5}
% & 5                       & 74.3\%          & 43.41\% &           
% \\ \hline
% \multirow{2}{*}{AGN-T}  & 2 & & &    \\ \cline{2-5}                
% & 5                       & 76.22\%         & 82.64\% &            
% \\ \hline
% \multirow{2}{*}{T+RL}     & 1           & 53.57\%         & 80.50\% & 66.70\%       \\ \cline{2-5} 
%                           & 1           & 76.96\%         & 85.50\% & 81.12\%      \\ \hline
% \multirow{2}{*}{AGN+RL}   & 1           & 53.31\%         & 62.50\% &  57.79\%     \\ \cline{2-5}                           & 1           & 76.23\%         & 76.04\% &  76.14\%     \\ \hline
% \multirow{2}{*}{AGN-T+RL} & 1           & 60.77\%         & 82.50\% &  71.37\%     \\ \cline{2-5}                           & 1           & 83.08\%         & 88.54\%  &  85.74\%    \\ \hline
% \end{tabular}\vspace{-0.5cm}
% \end{table}

\subsection{Use Case on Object Manipulation}
In this use case, it is assumed that the action sequences and task plans are already known in advance, but the desired manipulation locations for grasping and pushing need to be determined by the proposed framework.

\begin{figure}[!b]
    \centering\vspace{-0.5cm}
    \includegraphics[width=1.0\columnwidth]{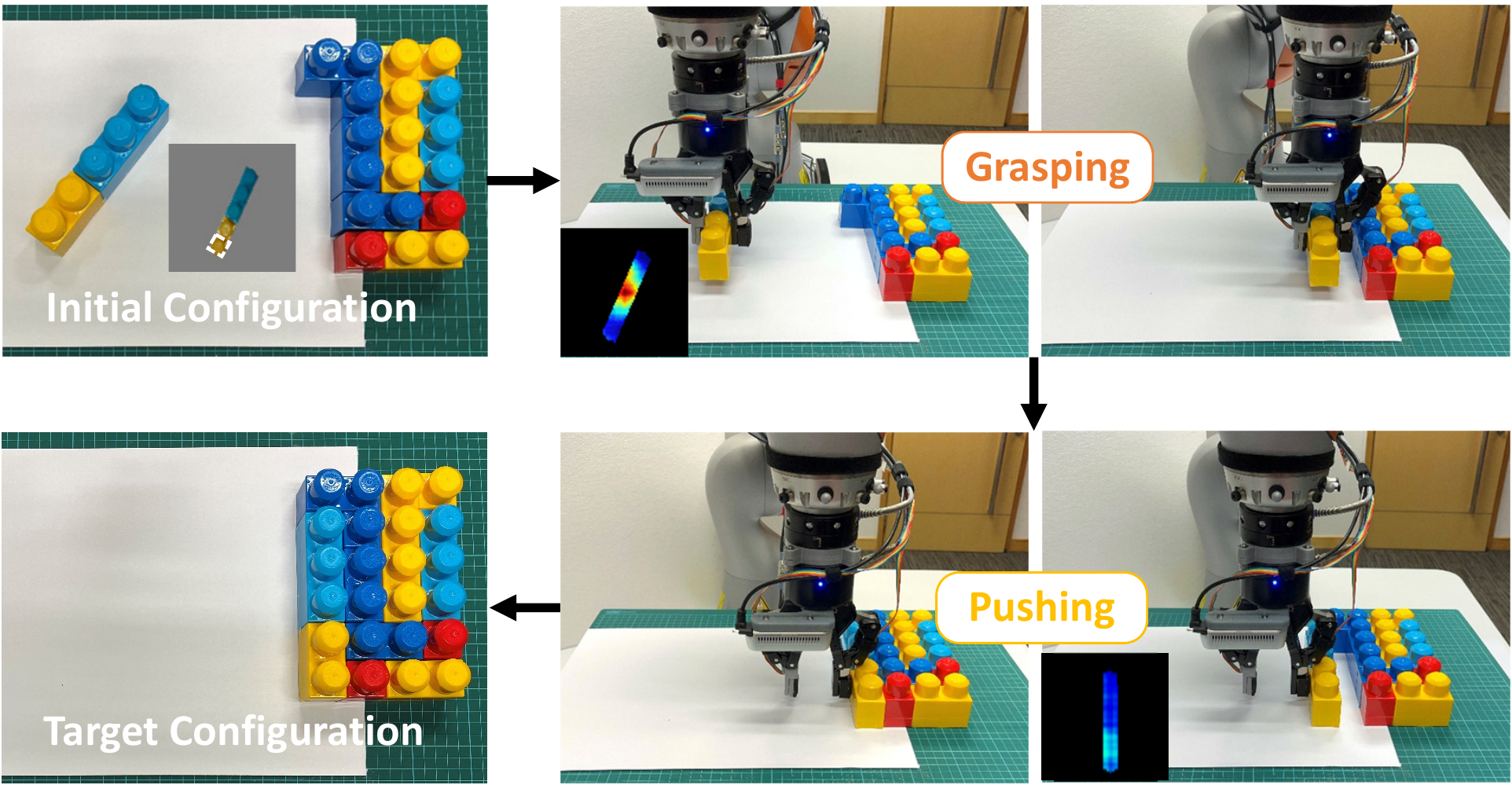}\vspace{-0.3cm}
    \caption{Robotic object manipulation with the proposed framework.}
    \label{fig:manipulation}
\end{figure}

Figure \ref{fig:manipulation} shows the sequence of actions for a successful manipulation of an object in a packing task. As can be seen, the object can first be grasped stably and then moved based on the affordance prediction model outputs and tactile feedback. Finally, it can be placed tightly with another object as expected (i.e., matches the target image) with pushing. 

\section{Conclusion}
In this paper, a multisensory-based object manipulation planning framework using multi-affordance model and reinforcement learning is developed to help the robot achieve different manipulation tasks on given objects with unknown intrinsic properties (e.g., CoM, mass distribution). The integrated vision and multimodal touch information with the attention mechanism is proposed to improve the accuracy and robustness of the affordance prediction model. To further improve the accuracy and efficiency, the learned affordance model is integrated into a DRL-based motion planning pipeline. The proposed framework is then tested on the YCBUSR dataset as well as our own dataset. The results on both datasets show that the proposed method achieves better accuracy, especially for the push affordance prediction. It can be also found from the results that the proposed framework can effectively achieve the manipulation tasks with high success rate and high efficiency.

%characteristics
% \addtolength{\textheight}{-11.3cm}
\bibliographystyle{IEEEtran}
\bibliography{ref.bib}

\end{document}